\documentclass[12pt]{article}
\usepackage{epsfig}
\usepackage{natbib}
\usepackage{multirow}
\usepackage{mathptmx,helvet,courier,footmisc,makeidx}
\usepackage{latexsym}
\usepackage{graphicx}
\usepackage{multicol}
\usepackage{makeidx}
\usepackage{url}
\usepackage{url}
\usepackage[colorlinks = true,
			linkcolor = blue,
			urlcolor  = blue,
			citecolor = blue,
			anchorcolor = blue]{hyperref}

\usepackage{soul} % So I can use \st for strikethrough
\usepackage{color}
\usepackage{palatino}
\usepackage{graphicx}

\newcommand{\E}{\mbox{E}}
\definecolor{eGreen}{rgb}{.057, .549,.065}

\usepackage{setspace}
\newcommand{\ybar}{\bar{y}}
\newcommand{\zbar}{\bar{z}}
\newcommand{\Ybar}{\bar{Y}}
\newcommand{\Ehat}{\hat{E}}
\newcommand{\Sigmahat}{\hat{\Sigma}}
\newcommand{\Sigmatilde}{\tilde{\Sigma}}
\newcommand{\Vtilde}{\tilde{V}}
\newcommand{\stilde}{\tilde{s}}
\newcommand{\yhat}{\hat{y}}
\newcommand{\alphahat}{\hat{\alpha}}
\newcommand{\gammahat}{\hat{\gamma}}
\newcommand{\etahat}{\hat{\eta}}
\newcommand{\muhat}{\hat{\mu}}
\newcommand{\deltahat}{\hat{\delta}}
\newcommand{\betahat}{\hat{\beta}}
\newcommand{\betatilde}{\tilde{\beta}}
\def\bmatrix#1{\left[ \matrix{#1} \right]}
\renewcommand{\E}{{\rm E}}
\newcommand{\SE}{{\rm SE}}
\newcommand{\Var}{{\rm Var}}
\newcommand{\Cov}{{\rm Cov}}
\newcommand{\tr}{{\rm tr}}
\newcommand{\vep}{\varepsilon}
\begin{document}

\title{Double Descent:  Understanding Linear Model Estimation of Nonidentifiable Parameters
and a Model for  Overfitting}

\author{Ronald Christensen  \\
Department of Mathematics and Statistics \\
University of New Mexico\footnote{Ronald Christensen is a Distinguished Professor in the Department of Mathematics and of Statistics, University of New Mexico, Albuquerque, NM, 87131.  Thanks to Mohammad Hattab for codesigning
and programing the simulation.}}

\date{August 26, 2024}
\maketitle

\begin{abstract}
We consider ordinary least squares estimation and variations on least squares estimation such as penalized
(regularized) least squares and spectral shrinkage estimates for problems with $p>n$ and associated problems
with prediction of new observations.  After the introduction of Section~1, Section~2 examines a number of
commonly used estimators for $p>n$.  Section~3 introduces prediction with $p>n$.
Section~4 introduces notational changes to facilitate discussion of overfitting and
Section~5 illustrates the phenomenon of double descent.  We conclude with some final comments.

\newpage
The original version of this \emph{ArXiv} article was submitted to \emph{The American Statistician}.
A revised version containing about half of this material was accepted for publication
under the title \emph{Linear Model Estimation and Prediction for $p > n$}.  The only
changes in this second \emph{ArXiv} version are to correct mistakes in the original that
were pointed out in the review process and to add a few clearly marked ``Later Notes.''\hfill
September 17, 2025
\end{abstract}

\setcounter{equation}{0}
\setcounter{page}{1}

\newpage

\section{Introduction}
Consider a standard linear model
\begin{equation}
Y=X\beta + e, \qquad \E(e)=0, \qquad \Cov(e)=\sigma^2 I_n.
\end{equation}
Here $Y$ is an observable random $n$ vector, $X$ is an observed $n \times p$ matrix, $\beta$ is a $p$ vector of unobservable fixed parameters, and $e$ is a unobservable
random vector of errors.  Linear models are extremely rich and include such nonparametric regression methods as fitting polynomials,
trigonometric functions, splines, wavelets, reproducing kernels, and fake reproducing kernels (like the hyperbolic tangent), cf. Christensen (2019, Chapter 1).
They easily incorporate generalized additive models.
Whenever $r(X)$, the rank of $X$, is less than $p$,
at least some of the parameters in $\beta$ are unidentifiable, thus making $\beta$ unidentifiable.
Of recent interest are problems in which $p > n$, which ensures that nonidentifiability exists in $\beta$.

We consider ordinary least squares estimation and variations on least squares estimation such as penalized
(regularized) least squares and spectral shrinkage estimates for problems with $p>n$ and associated problems
with prediction of new observations.  After the introduction of Section~1, Section~2 examines a number of
commonly used estimators for $p>n$.  Section~3 introduces prediction with $p>n$.
Section~4 introduces notational changes to facilitate discussion of overfitting and
Section~5 illustrates the phenomenon of double descent.  We conclude with some final comments.

When $p\leq n$, correct regression models should make valid predictions.  When overfitting occurs,
these predictions can often be improved using correct reduced models or biased estimation methods.
Letting $p \rightarrow n$ typically causes overfitting.  Because of identifiability issues,
there is no general guarantee that $p>n$ models, even when correct and with little variability $\sigma^2$,
will make valid predictions in the sense of predicting values close to the mean value of future observables.
However, there are situations in which good predictions are more likely to occur and, with randomly selected
training and test data sets, one can check how well particular procedures are working.
In many cases, notably nonparametric multiple regression, the many different approaches available
for modeling $E(Y)$ provide different possibilities for finding an approach that predicts well.

\subsection{Ordinary Least Squares}

Ordinary least squares (OLS) estimates minimize
    $$\|Y-X\beta\|^2 \equiv (Y-X\beta)'(Y-X\beta).$$
It is well known that OLS estimates are determined as solutions to either of the following equations
$$ X'X\beta = X'Y \qquad \hbox{or} \qquad X\beta = MY.$$
The first of these are the well known normal equations and can be obtained by setting derivatives equal to 0.
The second equation involves $M$, the unique (Euclidean) perpendicular projection operator (ppo) onto $C(X)$.
$C(X)$ is the column space of $X$, also known as the range space of $X$ and the span of the columns of $X$.
$M$ is unique, idempotent ($MM=M$), symmetric ($M=M'$), has $C(M)=C(X)$, and $MX=X$.
Christensen (2020, Chapter 2) [somewhat fancifully] refers to $\betahat$ being a least squares estimate if and only if it solves the second equation in the previous display as the Fundamental
Theorem of Least Squares Estimation.  When $r(X)< p$,
the least squares estimates $\betahat$ are not uniquely defined
but by the Fundamental Theorem and the uniqueness of $M$, the fitted values vector $X\betahat$ is unique.

\subsection{Penalized Least Squares}

Penalized (regularized) least squares estimates minimize, for some nonnegative penalty function $\mathcal{P}$ and nonnegative tuning parameter $\lambda$,
    $$\|Y-X\beta\|^2 + \lambda n \mathcal{P}(\beta).$$
Typically, $\mathcal{P}(0)=0$.  For any least squares estimates $\betahat$, uniquely define the sum of squares for error (SSE) as
    $$SSE \equiv \|Y-X\betahat\|^2$$
so that
    $$\|Y-X\beta\|^2 + \lambda n \mathcal{P}(\beta) = SSE + \|X\betahat-X\beta\|^2 + \lambda n \mathcal{P}(\beta).$$
From this second form it is clear that any penalized least squares estimates have to be functions of the OLS estimate $\betahat$ and indeed of the least squares fitted values $X\betahat$.  Typically, incorporating the penalty function is enough to make the penalized least squares estimates unique.

\subsection{Spectral Shrinkage Estimates}

Consider the singular value decomposition (SVD)
$$\frac{1}{n}X'X = VD(s_j)V'$$
wherein $D(s_j)$ is a $p \times p$ diagonal matrix of eigenvalues, the $j$th column of $V$ is an eigenvector of $\frac{1}{n}X'X$ corresponding to $s_j$,
and $V$ is an orthonormal matrix ($V'V=I_p=VV'$). Without loss of generality, assume $s_1 \geq s_2 \geq \cdots \geq s_p$.
With $r(X)=t$, we have $s_j>0$ for $j=1,\ldots,t$ and if $t<p$, then $s_j=0$ for $j=t+1,\ldots,p$.  In fact, with $V_r$ denoting the first $r$ columns of
$V$ and $D_r[\psi(s_j)]$ a diagonal matrix consisting of the function $\psi$ applied to the $r$ largest eigenvalues, another singular value decomposition is
    $$\frac{1}{n}X'X = V_tD_t(s_j)V_t'.$$
Here the columns of $V_t$ comprise an orthonormal basis for $C(X'X)=C(X')$ so that $N  \equiv V_tV_t' =  X'(XX')^-X$ is the
ppo onto $C(X')$, cf. Christensen (2020, Section~B.3).
When $p>>n$, the computational advantages become huge for performing an SVD on $\frac{1}{n}XX'$
with eigenvectors as columns
of a matrix $U$ and then computing $V$ from $U$, e.g. Christensen (2020, Section 13.3).

If $\Psi$ maps the nonnegative reals into themselves, define a \emph{spectral shrinker} as
$$\Psi\left( \frac{1}{n}X'X\right) \equiv VD\left[ \Psi(s_i)\right]V'.$$
The corresponding $\Psi$ shrinkage estimate is
\begin{eqnarray*}
\betahat_\Psi & \equiv & \Psi\left( \frac{1}{n}X'X\right) \frac{1}{n}X'Y \\
& = & \Psi\left( \frac{1}{n}X'X\right) \frac{1}{n}X'X\betahat,
\end{eqnarray*}
where the normal equations ensure that these are functions of the least squares estimates and fitted values.
We will see later that for spectral shrinkage estimates to actually shrink (a version of) the least squares estimates we need $\Psi(u)<1/u$.

\emph{[Later Note:  The use of $\Psi$ for both a function of matrices and of real numbers is
rather weird but it is not
a notation of my invention.  For ridge regression $\lambda$ is both of those functions as well as a real number.]}

\subsection{Identifiability}

Christensen (2020, Section 2.1) discusses identifiability in (generalized) linear models.
It amounts to the idea that identifiable functions are those that are functions of $X\beta$.
It is convenient to decompose $\beta$ into a component in $C(X')$ and a component perpendicular to $C(X')$
via $\beta=N\beta+(I-N)\beta$.  (Vectors $u$ and $w$ are perpendicular/orthogonal, denoted $u \perp w$, if and only
if $u'w=0$.  $C(X')^\perp$ is the vector space containing all vectors orthogonal to $C(X')$.)
Clearly, $N\beta = X'(XX')^-X\beta$ is a function of $X\beta$, hence identifiable and
therefore is something that can be estimated.
On the other hand, whenever $t<p$ so that $C(X') \neq \mathbf{R}^p$, neither $\beta$
nor the nontrivial parameter $(I-N)\beta$ is identifiable and cannot be uniquely estimated from data.
To see this consider two parameters $\beta_1$ and $\beta_2$ with
$\beta_1 \neq \beta_2$ but $N\beta_1 = N\beta_2$, then $X\beta_1=X\beta_2$, so any functions of $X\beta$ must be the same without the parameters being the same, hence
$\beta$ is not a function of $X\beta$ and not identifiable.   Similarly, under the same conditions, while $X\beta_1=X\beta_2$, the function $(I-N)\beta$ displays $(I-N) \beta_1 \neq (I-N)\beta_2$, so $(I-N)\beta$ is not
identifiable.

The distribution of the data $Y$ depends only on $N\beta$.
The parameter $(I-N)\beta$ has no effect on the distribution of $Y$,
so there is no information in the data about $(I-N)\beta$.
Even a Bayesian who puts a prior distribution on $\beta$ thereby determines a conditional distribution for
$(I-N)\beta$ given $N\beta$ and this conditional distribution will be the same in the prior and the posterior.
Moreover, the distribution of any future observation will depend on its mean,
and if that mean is not identifiable, it will depend on $(I-N)\beta$ which we have no way to learn about except
via prior information.

\setcounter{equation}{0}
\section{Estimation in Nonidentifiable Models}

\emph{[Later Note:  A referee pointed out how bad the original section title was.]}

Commonly used methods for estimating nonidentifiable linear model parameters include penalized least squares, spectral shrinkage estimates, and
the minimum norm (shortest) least squares estimate, see Hastie et al.~(2020) and Richards et al.~(2020).
We just saw that all penalized least squares and spectral shrinkage estimates are functions of the OLS estimates, indeed of the unique least squares fitted values.
It follows immediately that they will, in particular, be functions of the minimum norm least squares estimates.

In this section we discuss ridge regression, minimum norm least squares, gradient descent estimates, and principal component regression estimates.
Along the way we provide simple proofs of some well known facts: that minimum norm least squares estimates are unique,
that when properly initialized, a convergence point of gradient descent must be the minimum norm OLS estimate,
that the Moore-Penrose generalized inverse OLS estimate gives the minimum norm OLS estimate,
that letting the ridge parameter go to zero gives minimum norm OLS, and that using all of the
worthwhile principal components gives minimum norm OLS.

Ridge regression estimates were all the rage in the 1970s.
They then went out of fashion for a long time but have now experienced a rebirth with the existence of large data sets.
Ridge regression estimates are defined as penalized least squares estimates,
$$\betahat_\lambda \equiv \arg\min_\beta \left\{ \frac{1}{n} \| Y- X \beta\|^2 + \lambda \| \beta \|^2 \right\}.$$
Note that this penalty function does not make a lot of sense unless the columns of $X$ are on some common scale, a statement
that also applies to most other off-the-shelf penalty functions like LASSO and elastic net.

Ridge estimates have a well known alternative formulation in terms of a matrix formula used below.  We now show the unsurprising fact that
ridge estimates are also spectral shrinkage estimates.   Choose the spectral shrinker function
$$\lambda(u) \equiv \frac{1}{u+\lambda}.$$
The scalar $\lambda$ determines the function $\lambda$ and vice versa.
Write the ridge regression estimate as
\begin{eqnarray*}
\betahat_\lambda & \equiv & (X'X + \lambda n I_p)^{-1} X'Y \\
    & = &  \left( \frac{1}{n}X'X + \lambda I_p \right)^{-1} \frac{1}{n}X'Y \\
    & = &  \left[ VD(s_j)V' + \lambda V V' \right]^{-1} \frac{1}{n}X'Y \\
    & = &  \left[ VD(s_j + \lambda ) V' \right]^{-1} \frac{1}{n}X'Y \\
    & = & VD[1/(s_j +\lambda)]V' \frac{1}{n}X'Y \\
    & \equiv & \lambda\left( \frac{1}{n}X'X \right) \frac{1}{n}X'Y .
\end{eqnarray*}
This also establishes that for any $\lambda >0$, the inverse $(X'X+\lambda n I)^{-1}=VD[n/(s_j+\lambda)]V'$ always exists so that ridge estimates are always unique.

Ridge estimators are often compared to minimum norm least squares estimates, that is,
the least squares estimate $\betahat$ with the smallest value of $\|\betahat\|^2 \equiv \betahat'\betahat$.
Any OLS estimate $\betahat$ has $\betahat = N \betahat + (I-N)\betahat$ and
$\|\betahat\|^2 =  \|N\betahat\|^2 + \|(I-N)\betahat\|^2$.
Since $NX'=X'$, we get $XN=X$ and $X\betahat = X(N\betahat)$ so, by the Fundamental Theorem of Least Squares Estimation,
$N\betahat$ is an OLS estimate that is no longer than $\betahat$.

The search for minimum norm least squares estimates can now be restricted to those in $C(X')$.
If there is a unique estimate in $C(X')$ we are done.  As argued in Nosedal-Sanchez et al.~(2012)
and in Christensen (2019, Section 3.1),
if $\betahat$ and $\betatilde$ are both OLS estimates
in $C(X')$, then $(\betahat - \betatilde) \in C(X')$ but also from the Fundamental Theorem,
$X(\betahat - \betatilde)=0$, so $(\betahat - \betatilde) \perp C(X')$,
i.e. $(\betahat - \betatilde) \in C(X')^\perp$.
But the intersection of the two spaces $C(X')$ and $ C(X')^\perp$
is the zero vector because the only vector orthogonal to itself is the zero vector.  Hence $\betahat = \betatilde$.

Denote the minimum norm OLS estimate $\betahat_m$.  For any OLS $\betahat$ we have $\betahat_m=N\betahat$.  Another method of finding
$\betahat_m$ is to fit the model $Y=XX'\delta+e$ by ordinary least squares and take $\betahat_m=X'\deltahat$.  This works because $C(X)=C(XX')$, so they have
the same ppo $M$ and by the Fundamental Theorem applied to this and model (1.1), $X\betahat=MY=XX'\deltahat$,
so $X'\deltahat$ is an OLS estimate of $\beta$ but is obviously in $C(X')$.
For any generalized inverse of $X'X$, say $(X'X)^-$, it is well known that $\betahat = (X'X)^-X'Y$ is an OLS estimate.
The Moore-Penrose generalized inverse of $X'X$, say $(X'X)^+ \equiv V_tD_t(/ns_j)V_t'$, has the property that $C[(X'X)^+]=C(V_t)=C(X')$.
Therefore, the Moore-Penrose OLS estimate $\betahat^+ \equiv (X'X)^+X'Y \in C(X')$ must be the minimum norm least squares estimate.
Defining the spectral shrinker function $\xi(u) \equiv \cases{ 1/u & if $u>0$ \cr 0 & if $u=0$}$, the Moore-Penrose estimate is also the spectral shrinkage estimate
    $$\betahat_\xi = \betahat^+ = \betahat_m.$$

We now show the often stated but rarely proven fact that, if the gradient (steepest) descent algorithm is properly initialized and converges,
it must be to the minimum norm least squares estimate $\betahat_m$.  Set up gradient descent as minimizing $\|Y-X\beta \|^2$ initialized at
$\beta_0$ with step size $\eta/n>0$, then
$$\beta_{t+1} \equiv \beta_t + \frac{\eta}{n} X'(Y-X\beta_t).$$
Clearly, any convergence point ($\beta_{t+1}=\beta_t$) satisfies the normal equations ($X'Y=X'X\beta_t$) hence is a least squares estimate.
Generally, if $\beta_0 \in C(X')$, then all $\beta_t$s are in $C(X')$ and any convergence point has to be the unique least squares estimate in $C(X')$.
So if  $\beta_0 \in C(X')$, gradient descent can only converge at the minimum norm least
squares estimate $\betahat_m$.  Obviously, the common choice $\beta_0=0 \in C(X')$.  For a completely general $\beta_0$, a point of convergence would be the OLS estimate $\betahat_m+(I-N)\beta_0$.
If you had prior information on $\beta$, which seems like it would be difficult to elicit when $p>n$, you could start gradient descent at your best prior guess and likely get a reasonable
OLS estimate that is not in $C(X')$.

Richards et al.~(2021) argue that the entire sequence of gradient descent iterations are interesting spectral
shrinkage estimates with $\beta_t$ defined by applying the spectral shrinker function
$$t(u) \equiv \eta\sum_{r=0}^{t-1}(1-\eta u)^r = (1-(1-\eta u)^t)/u.$$
For this to actually shrink we need $\eta \max_j(s_j) <1$.

The simplest and most familiar model in which the parameters are nonidentifiable is the balanced one-way analysis of variance,
    $$y_{ij} = \mu + \alpha_i + \vep_{ij}, \quad i=1,\ldots, a, \quad j=1,\ldots,N.$$
For computer programs to print out a table of estimated coefficients, they need to select a choice of least squares estimates.  Corresponding to four common
side conditions, four common choices of OLS estimates are
$$ \betahat_1 = \bmatrix{0 \cr \ybar_{1\cdot} \cr \ybar_{2\cdot} \cr \vdots \cr \ybar_{a\cdot}} \qquad   \betahat_2 = \bmatrix{\ybar_{\cdot \cdot} \cr \ybar_{1\cdot} - \ybar_{\cdot \cdot}  \cr \ybar_{2\cdot} - \ybar_{\cdot \cdot} \cr \vdots  \cr \ybar_{a\cdot} - \ybar_{\cdot \cdot} } \qquad %$$ $$
\betahat_3 = \bmatrix{ \ybar_{1\cdot} \cr 0 \cr \ybar_{2\cdot} - \ybar_{1\cdot} \cr \vdots \cr \ybar_{a\cdot} - \ybar_{1\cdot}} \qquad \betahat_4 = \bmatrix{ \ybar_{a \cdot} \cr  \ybar_{1 \cdot} -\ybar_{a \cdot} \cr \vdots \cr \ybar_{a-1\cdot} - \ybar_{a\cdot}\cr 0} . $$
As shown in the appendix, the minimum norm OLS and ridge estimators are
$$\betahat_m = \bmatrix{ \frac{a}{a+1}\ybar_{\cdot \cdot} \cr \ybar_{1 \cdot} - \frac{a}{a+1}\ybar_{\cdot \cdot} \cr \vdots \cr  \ybar_{a \cdot} - \frac{a}{a+1}\ybar_{\cdot \cdot} }; \qquad \betahat_\lambda = \bmatrix{\frac{a}{a+1+a\lambda} \ybar_{\cdot \cdot}  \cr \frac{1}{1+a\lambda}\left(\ybar_{1 \cdot} -  \frac{a}{a+1+a\lambda} \ybar_{\cdot \cdot} \right)\cr \vdots \cr \frac{1}{1+a\lambda} \left(\ybar_{a \cdot} -  \frac{a}{a+1+a\lambda} \ybar_{\cdot \cdot} \right)}.$$

The balanced additive two-way ANOVA is
    $$y_{ijk} = \mu + \alpha_i + \eta_j + \vep_{ijk}, \quad i=1,\ldots, a, \quad j=1,\ldots,b, \quad k=1,\ldots,N.$$
Letting $J_r$ denote an $r$ vector of 1s, one well-known choice of least squares estimates is
$$ \betahat \equiv \bmatrix{\muhat \cr \alphahat \cr \etahat} = \bmatrix{\ybar_{\cdot \cdot \cdot} \cr \Ybar_a - \ybar_{\cdot \cdot \cdot}\, J_a \cr \Ybar_{b} - \ybar_{\cdot \cdot \cdot}\, J_b },
\quad \hbox{where} \quad \Ybar_{a}= \bmatrix{\ybar_{1\cdot\cdot} \cr \ybar_{2\cdot \cdot} \cr \vdots \cr \ybar_{a \cdot\cdot} };
\quad  \Ybar_{b}= \bmatrix{\ybar_{\cdot 1 \cdot} \cr \ybar_{\cdot 2 \cdot} \cr \vdots \cr \ybar_{\cdot b \cdot} }.
$$
Note that $J_a'\Ybar_a=a\ybar_{\cdots}$ and $J_b'\Ybar_b=b\ybar_{\cdots}$.
As shown in the appendix,
$$\betahat_m \equiv \bmatrix{\muhat_m \cr \alphahat_m \cr \etahat_m} =
\bmatrix{\frac{ab}{a+b+ab}\ybar_{\cdot \cdot \cdot} \cr \Ybar_a - \frac{a+ab}{a+b+ab}\ybar_{\cdot \cdot \cdot}\, J_a \cr
\Ybar_{b} - \frac{b+ab}{a+b+ab}\ybar_{\cdot \cdot \cdot}\, J_b } .$$
Note that $\betahat_m$ is a least squares estimate because
$\yhat_{ijk} = \ybar_{i \cdot \cdot} + \ybar_{\cdot j \cdot} - \ybar_{\cdot \cdot \cdot} = \muhat_m + \alphahat_{mi}+\etahat_{mj}$.

Consider again a general spectral shrinkage estimate.  We now show that all spectral shrinkage estimates have $\betahat_\Psi \in C(X')$.
The normal equations give $X'Y = X'X\betahat_m$ and, recalling that $s_j>0$ for $j=1,\ldots, t$ and when $t<p$, $s_j=0$ for $j=t+1,\ldots, p$, we obtain
\begin{eqnarray*}
\betahat_\Psi & \equiv & \Psi\left( \frac{1}{n}X'X\right)^{-1} \frac{1}{n}X'Y \\
& = & VD\left[\Psi(s_j)\right]V' \frac{1}{n}X'X \betahat_m  \\
& = & VD\left[\Psi(s_j)\right]V' V D(s_j)V' \betahat_m  \\
& = & VD\left[\Psi(s_j)\right]D(s_j)V' \betahat_m  \\
& = & VD\left[\Psi(s_j)s_j\right]V' \betahat_m  \\
& = & V_tD_t\left[\Psi(s_j)s_j\right]V_t' \betahat_m .
\end{eqnarray*}
Thus spectral shrinkage estimates are always in $C(X')$.
If $\Psi(s_j)s_j=1$, $j=1,\ldots t$, then $\betahat_\Psi=\betahat_m$,
so for spectral shrinkage estimates to shrink the least squares estimates, we want $\Psi(u)<1/u$.
Such spectral shrinkage estimates are shrinking the components of the vector $\gammahat_t = V_t'\betahat_m$ that we will soon see to be the OLS estimate of
the regression coefficients $\gamma_t$ in the principal component regression model $Y=(XV_t)\gamma_t+e$.  Incidentally, generalized ridge regression using the
penalty function $\beta'Q\beta$ for positive definite $Q$ does not seem to keep the estimates in $C(X')$ unless $C(QX') = C(X')$.

For standard ridge regression the previous paragraph gives
    $$\betahat_\lambda = V_tD_t[s_j/(s_j+\lambda)]V_t'\betahat_m .$$
It is now immediate that as the ridge parameter $\lambda$ converges to 0, the ridge estimate converges to the minimum norm OLS estimate.
As $\lambda$ converges to 0, $s_j/(s_j+\lambda)$ converges to 1 for $j=1,\ldots,t$, so $V_tD_t[s_j/(s_j+\lambda)]V_t'$ converges to $V_tV_t'=N$
and $\betahat_\lambda$ converges to $N\betahat_m = \betahat_m$, the unique minimum norm OLS estimate.
Moreover, the ridge estimates only substantially differ from the minimum norm OLS estimates in eigenvector directions for which
the eigenvalues are small.  When $s_j$ is large, $s_j/(s_j+\lambda)$ is close to one and when $s_j$ is small, $s_j/(s_j+\lambda)$ is close to zero.
To the extent that we can categorize all of the $s_j$ values as either large or small, we will see shortly that the ridge estimates approximate principal component regression.
Recall that, ignoring the (critical) issues of location adjustment and scaling, the columns of
$XV_t$ constitute the $t$ nontrivial principal components associated with $X'X$.

Another method used when $p > n$ is principal component regression, e.g. Hattab, Jackson, and Huerta (2019).
Again ignoring the issues of location adjustment and scaling of the predictor variables,
regression on the first $r\leq t$ principal components fits the model $Y=(XV_r) \gamma_r + e$ by OLS.
The model implicitly presupposes that $\beta = V_r\gamma_r \in C(V_t) =C(X')$, so the estimate of $\beta$ becomes $\betahat_{Pr} \equiv V_r \gammahat_r$ or,
\begin{eqnarray*}
\betahat_{Pr} & = & V_r[V_r'X'XV_r]^{-1}V_r'X'Y \\
    & = & V_r[V_r'X'XV_r]^{-1}V_r'X'X\betahat \\
    & = & V_r[V_r'V_tD_t(ns_j)V'V_r]^{-1}V_r'V_tD_t(ns_j)V_t'\betahat \\
    & = & V_r\left\{[I_r,0]D_t(ns_j)\bmatrix{I_r \cr 0}\right\}^{-1}[I_r,0]D_t(ns_j)V_t'\betahat \\
    & = & V_r\left\{D_r(ns_j)\right\}^{-1}D_r(ns_j)V_r'\betahat \\
    & = & V_rV_r'\betahat \\
    & = & V_rV_r' N \betahat \\
    & = & V_rV_r' \betahat_m
\end{eqnarray*}
which projects $\betahat_m$ into the space spanned by the first $r$ eigenvectors of $X'X$.
Because it is a projection of the minimum norm OLS estimate, $\betahat_{Pr}$ must also be unique.
This seems not quite to qualify as a spectral shrinkage estimate because the necessary $\Psi$ function would need
to depend on the specific eigenvalues of $\frac{1}{n}X'X$.  If, rather than specifying the number of principal components $r$, you specify how large
the eigenvalue has to be to be included, that would be a spectral shrinkage estimate.  Christensen (2020, Section 13.1) argues that principal
components corresponding to small eigenvalues should be excluded from the analysis because they are unreliable directions in the estimation space.
Marquardt's generalized inverse regression is equivalent to principal component regression, again ignoring issues of location adjustment and scaling.

Note that $C(XV_t)=C(X)$, so the $t$ dimensional full model  $Y=(XV_t) \gamma_t + e$ is equivalent to the original linear model but
reparameterizes it so that the unique estimate $\gammahat_t=V_t'\betahat_m$ determines the minimum norm OLS estimate
$\betahat_{Pt} \equiv V_t\gammahat_t=V_tV_t'\betahat_m =\betahat_m$. The variance-bias trade-off discussion
of Christensen (2020, Section 2.9) arguing that correct reduced models always give improved estimation and that even incorrect reduced models often improve estimation
applies directly to the full model $Y=(XV_t) \gamma_t + e$ and reduced models $Y=(XV_r) \gamma_r + e$.
%Note that what spectral shrinkage estimates are shrinking are the estimates  $V_t'\gammahat_t$ of the regression coefficients in principal component regression $V_t'\gamma_t$.
In particular, when the reduced model is (close to) correct, the OLS estimate $\gammahat_r$ should be an improvement over the OLS $\gammahat_t$.
Typically  $Y=(XV_r) \gamma_r + e$ is an incorrect model, so $\gammahat_r$ typically provides a biased estimate of $\gamma_t$ but may have smaller overall error than $\gammahat_t$.
These properties extend to $\betahat_{Pr} \equiv V_r \gammahat_r$ being perhaps biased for $N\beta=NV_t\gamma_t$ but perhaps better overall than $\betahat_{Pt}=V_t\gammahat_r=\betahat_m$.

Finally, consider the LASSO.  It is difficult to examine the LASSO applied directly to $\beta$ but we consider a transformed LASSO.
As discussed earlier, any penalized (regularized) least squares estimate can be found by minimizing
$$\frac{1}{n}\|X\betahat_m-X\beta\|^2 +\lambda \mathcal{P}(\beta)= \frac{1}{n}(\betahat_m-\beta)'X'X (\betahat_m-\beta)+\lambda \mathcal{P}(\beta).$$
Suppose we specify a penalty function of the form
    $$\mathcal{P}(\beta) = \sum_{j=1}^p \psi(e_j'V'\beta)$$
where $\psi(u)\geq 0$, $\psi(0)=0$, and $e_j$ is a $p$-vector of 0s except for a 1 in the $j$th spot.
If we now reparameterize and also transform the estimates by
$$\gamma \equiv V'\beta; \qquad \gammahat_m \equiv V'\betahat_m,$$
the function we need to minimize becomes
\begin{equation}
(\gammahat_m-\gamma)'D(s_j)(\gammahat_m-\gamma)+ \lambda \sum_j \psi(\gamma_{j}) = \sum_{j=1}^p \left[ s_j (\gammahat_{mj}-\gamma_j)^2 + \lambda \psi(\gamma_j)\right].
\end{equation}
The advantage of this form is that we can minimize the overall function by minimizing each of the $p$ component functions.
For any $j$ with $s_j=0$, the function will be minimized by taking $\gamma_j=0$ and we only need to minimize
     $$\sum_{j=1}^t \left[ s_j (\gammahat_j-\gamma_j)^2 + \lambda \psi(\gamma_j)\right].$$
Arguments similar to those in Christensen (2019, Section 2.3) show that, with the LASSO penalty $\psi(u)=|u|$, the estimate $\gammahat_L$ is determined by
$\gammahat_{Lj}=0$, for $j=t+1,\ldots,p$ and for $j=1,\ldots,t$,
    $$\gammahat_{Lj}= \cases{
\gammahat_{mj} - \lambda/2s_j & if $\gammahat_{mj} \geq \lambda/2s_j$ \cr
 0  & if $|\gammahat_{mj}| < \lambda/2s_j$ \cr
\gammahat_{mj} + \lambda/2s_j & if $\gammahat_{mj} \leq -\lambda/2s_j$
}$$
Reverting back to the original parameterization, $\betahat_L \equiv V\gammahat_L$ and, because of $\gammahat_{Lj}=0$ for $j=t+1,\ldots,p$,
we have $\betahat_L \in C(V_t)=C(X')$.  This LASSO performs both variable selection and shrinkage estimation on the principal components regression model.

In general, if we use $\gammahat_{\stackrel{\bullet}{\psi}}$ to denote the minimizer of (\thesection.1) and denote the derivative of $\psi(u)$ as $\stackrel{\bullet}{\psi}(u)\equiv \mathbf{d}_u \psi(u)$,
then $\gammahat_{\stackrel{\bullet}{\psi} j}=0$ for $j=t+1,\ldots,p$ and for $j \leq t$ the estimate  $\gammahat_{\stackrel{\bullet}{\psi} j}$ will take on one of the $u$ values at which
the derivative does not exist or it will be a solution to $\gammahat_{mj} = \gamma_j - (\lambda/2s_j)\stackrel{\bullet}{\psi}(\gamma_j)$.
We just need to check which of these values minimize (\thesection.1).  In any case, $\betahat_{\stackrel{\bullet}{\psi}} \equiv V\gammahat_{\stackrel{\bullet}{\psi}}$ will be in $C(V_t)=C(X')$ because
$\gammahat_{\stackrel{\bullet}{\psi} j}=0$ for $j=t+1,\ldots,p$.

\subsection{Will these work?}

In linear model theory it is common to be concerned with estimating linear functions of $\beta$, say
$\Lambda'\beta$ for some known $p \times r$ matrix $\Lambda$.
As we will see, the standard problem of predicting
future observations is a special case of this estimation problem.  In traditional $p \leq n$ linear models
there has been little need to consider estimation of any such functions that are not identifiable.
Such functions are said to be ``estimable'' if there exists a matrix $P$ such that $\Lambda'=P'X$.
With this structure, $\Lambda'\beta= P'X\beta$ is clearly a function of $X\beta$ and is identifiable.
Moreover, $\Lambda'\beta= P'X\beta =  P'XN\beta =\Lambda'(N\beta)$,
so estimation involves $N\beta$ but never involves the unidentifiable parameter $(I-N)\beta$.

The estimation methods we have considered revolve around the minimum norm OLS estimate $\betahat_m$ which provides an unbiased estimate of $N \beta$.
To see unbiasedness note that $N=X'(XX')^-X$ and by the Fundamental Theorem, for any OLS estimate
    $\E(X\betahat)=\E(MY) = M\E(Y) = MX\beta  = X\beta,$
so in particular we have
    $$\E(\betahat_m)=\E(N\betahat)=\E[X'(XX')^-X\betahat]=X'(XX')^-\E(X\betahat)=X'(XX')^-X\beta =N\beta.$$
We can possibly improve on this estimate by fitting reduced models or using
other forms of shrinkage estimation as discussed earlier.

Unfortunately, with $p>n$ restricting $\Lambda'\beta$ to estimable functions seems to be
untenable, especially when $\Lambda'\beta$ is taken to be the mean of some future observations.
\textbf{In problems for which $\Lambda'(I-N)\beta$ is substantially different from 0 there is simply no
hope of ensuring that our estimation methods will do a good job.}
We might be able to use prior information on $\beta$ to inform our decisions on $(I-N)\beta$ but
even with $p \leq n$ it is difficult to elicit meaningful prior information on $\beta$
and with $p > n$ it is well-nigh impossible.  \emph{The remainder of this
discussion looks for situations in which $\Lambda'\beta$ is somehow close to being estimable
so that it is plausible that  $\Lambda'(I-N)\beta$ is not substantially different from 0 and
estimation methods based on $\betahat_m$ may do a good job.}

\setcounter{equation}{0}
\section{Prediction}
Typically, we want to use model (1.1) to predict the observations $Y_f$
in a future $n_f$ dimensional linear model with the same $\beta$ parameter,
\begin{equation}
Y_f = X_f \beta + e_f.
\end{equation}
Model (1.1) was taken as a standard model with mean 0, homoscedastic, uncorrelated -- now independent -- errors
and the same assumptions are typically made about the predictive model.
Also $Y_f$ and $Y$ are assumed independent.
When $p \leq n$ it is standard practice (cf., Christensen, 2020, Section~6.6 or Christensen, 2024)
to assume that $X_f\beta$ is estimable in model (1.1).
\emph{Predictive estimability} means that there exists a matrix
$P_f$ such that $X_f = P_f'X$, which implies that $C(X_f')=C(X' P_f) \subset C(X')$.  It follows that
for these prediction problems $X_f(I-N)\beta=0$ and we do not need to worry about $(I-N)\beta$.
With $p<n$, predictive estimability automatically holds for all regression models ($t=p$) and
does not seem like an undue restriction for models with categorical variables,
but requiring predictive estimability seems overly restrictive when considering models with $p > n$.

It is easily seen that any estimate of $\beta$ from model (1.1), say $\betatilde(Y)$, has $Y_f$ and $\betatilde(Y)$ independent, so the expected squared prediction error satisfies
    $$\E \| Y_f - X_f\betatilde(Y)\|^2 = \E \| Y_f - X_f\beta\|^2 + \E \| X_f\beta - X_f\betatilde(Y)\|^2 .$$
Thus the prediction problem reduces to the problem of estimating $X_f\beta$.
If we knew what $\beta$ was, the best possible predictor of $Y_f$ would be $X_f\beta$.

\subsection{Partitioned Models}

Nobody uses the earlier theory directly, it needs to be adapted to partitioned models.  Christensen (2019, Chapter~2) discusses penalized least squares for general partitioned models but here
we restrict ourselves to the simplest and most common case.  Suppose the vector of predictor variables $x$ always has a lead element of 1, so that $x'=(1,z')$.
Write $$X = \bmatrix{x_1' \cr \vdots \cr x_n'}= \bmatrix{1 & z_1' \cr \vdots & \vdots \cr 1 & z_n'} = \left[J_n, Z\right].$$
Model (1.1) becomes
        $$Y = J_n \beta_0 + Z\beta_* + e.$$
Writing $J_r^r \equiv J_rJ_r'$, the model can be reparameterized as
\begin{equation}
Y = J_n \alpha + [I-(1/n)J_n^n]Z \beta_* + e.
\end{equation}
Note that both $(1/n)J_n^n$ and $I-(1/n)J_n^n$ are ppos. $[I-(1/n)J_n^n]Z=Z-J_n\zbar_\cdot'$ has each column in $Z$ corrected for its sample mean value.  (More often than not the columns would also be scaled to have a common length.)

It is well known that the OLS estimate of $\alpha$ is $\ybar_\cdot$, the sample mean.  OLS estimates of $\beta_*$ are obtained by solving the normal equations or
the Fundamental Theorem equations
    $$Z'[I-(1/n)J_n^n]Z \beta_*  = Z'[I-(1/n)J_n^n]Y; \qquad [I-(1/n)J_n^n]Z \beta_* = M_*Y,$$
where $M_*$ is the ppo onto $C\left\{[I-(1/n)J_n^n]Z \right\}$.
Define
    $$S_{zz} \equiv \frac{1}{n-1} Z'[I-(1/n)J_n^n]Z; \qquad S_{zy} \equiv \frac{1}{n-1} Z'[I-(1/n)J_n^n]Y.$$
Note that the normal equations can now be rewritten as
    $$ S_{zz}\beta_* = S_{zy}.$$
When $(y_i,z_i')'$ form a random sample from some population, $S_{zz}$ and $S_{zy}$ are the
standard unbiased estimates of $\Cov(z) \equiv \Sigma_{zz}$ and $\Cov(z,y) \equiv \Sigma_{zy}$.
Penalty functions typically only involve $\beta_*$, so are written $\mathcal{P}(\beta_*)$ and penalized least squares estimates are found by minimizing, for any OLS
estimate $\betahat_*$,
    $$\frac{1}{n-1}(\betahat_* - \beta_*)'\Big\{ Z'[I-(1/n)J_n^n]Z\Big\}(\betahat_* - \beta_*)  + \lambda \mathcal{P}(\beta_*).$$
Spectral shrinkage estimates involve the SVD of $S_{zz}$,
    $$S_{zz}=VD(s_j)V'$$
and
    $$\betahat_{*\Psi} \equiv \Psi(S_{zz})S_{zy}.$$
With $N_*$ the ppo onto $C\left\{Z'[I-(1/n)J_n^n] \right\} = C\left\{Z'[I-(1/n)J_n^n]Z \right\}$, the relevant decomposition of $\beta_*$
into identifiable and nonidentifiable components is $\beta_* = N_* \beta_* + (I-N_*) \beta_*$.

The best possible prediction from this model for a future case $y_{fi}$ with predictor variables (features) $z_{fi}$ is  $\E(y_{fi})=\alpha + (z_{fi}-\zbar_\cdot)'\beta_*$.
The OLS estimate of this is
    $$\yhat_{fi} \equiv \ybar_\cdot + (z_{fi}-\zbar_\cdot)'\betahat_*.$$
Better predictions may or may not be obtained by replacing $\betahat_*$ with a penalized estimate or a spectral
shrinkage estimate or a principal components estimate.
For the parameter $\alpha + (z_{fi}-\zbar_\cdot)'\beta_*$ to be estimable we need $(z_{fi}-\zbar_\cdot) \in C(S_{zz}) = C\{Z'[I-(1/n)J_n^n]\}$.

The reparameterized version of the partitioned predictive model is
    $$Y_f=J_{n_f}\alpha + [Z_f- J_{n_f}\zbar_\cdot']\beta_* + e_f,$$
where $\zbar_\cdot'=(1/n)J_n' Z$ so that the columns of $Z_f$ are location corrected by data
from model (1.1) rather than using $[I_{n_f}-(1/n_f)J_{n_f}^{n_f}]Z_f$.

\emph{It is easy to tell if any future case is a good candidate for making a prediction.
Simply regress $z_{fi}-\zbar_\cdot$ on
the columns of $Z'[I-(1/n)J_n^n]$.  If this fit has the sum of squared errors close to 0,
the predictive case is close to being estimable
and is less likely to be thrown off by the nonidentifable quantity $(z_{fi}-\zbar_\cdot)'(I-N_*)\beta_*$.
What we would like is some assurance that all of the cases at which we might want to predict will be
good candidates for making a prediction.}

Fitting high dimensional polynomials is notorious as a method for overfitting data.  Suppose we have a univariate predictor $w$.  With 7 distinct $w$ values we can fit a 6th degree polynomial which has $p=7$.
However, suppose the
$w$ values form 3 widely separated clusters, say $w \in \{0, 0.5, 1, 10, 19, 19.5, 20\}$.
One cannot meaningfully fit anything more than a parabola to such data and fitting 4th, 5th, or 6th
degree polynomials would be considered overfitting and can give wild predictions for
$w$ values between the clusters.
Fortunately it is easy to see that trying to predict at, say, $w=5$ using a high degree polynomial
would be a poor idea because the
\emph{predictive leverage}, defined as $\Var(x_f'\betahat)/\sigma^2 = [\SE(x_f'\betahat)]^2/MSE$, will be large,
cf.~Christensen (2015, Section 9.2).

The predictive estimability issue is something different.
To get good predictions we will always need $n$ somewhat large but for illustration consider $n=3$, $p-1=4$, and $z=(w,w^2,w^3,w^4)$ with $w=-1,0,1$.  For predictions consider $z_f$ based on $w_f=0.5,1.5$.  It is not to hard to see that $$Z= \bmatrix{-1 & 1 & -1 & 1 \cr 0 & 0 & 0 & 0 \cr 1 & 1 & 1 & 1 }
\qquad \hbox{and} \qquad C\left\{ Z'[I-(1/n)J_n^n] \right\} = C\bmatrix{1 & 0 \cr 0 & 1 \cr 1 & 0 \cr  0 & 1 \cr   }. $$
Obviously, we can get good predictions whenever $w_f$ is close to any of $-1,0,1$ but for the relatively difficult predictions at $w_f=0.5,1.5$,
$$Z_f'-\zbar_\cdot J_2'= \bmatrix{1/2 & 3/2 \cr -5/12 & 19/12 \cr 1/8 & 27/8 \cr -29/48 & 95/48} \doteq
\bmatrix{0.5 & 1.5 \cr -0.4 & 1.6 \cr 0.1 & 3.4 \cr -0.6 & 2.0}.$$
While neither of the vectors $z_{fi}-\zbar_\cdot$ is all that close to being in
$C\left\{ Z'[I-(1/n)J_n^n] \right\}$, clearly the first column $z_{f1}-\zbar_\cdot$, corresponding to $w_f=0.5$,
is much closer to predictive estimability than is $z_{f2}-\zbar_\cdot$ with $w_f=1.5$.
Of course what we really need is to have $(z_{fi}-\zbar_\cdot)'(I-N_*)\beta_* \doteq 0$ but in the absence of
knowledge about $(I-N_*)\beta_*$, all we can do is hope that having
$(z_{fi}-\zbar_\cdot)'(I-N_*) \doteq 0$ will yield good predictions.
Of course in this example if $n\geq 5$ with at least 5 distinct $w$ values,
predictive estimability will always hold exactly for the 4th degree polynomial,
regardless of issues of overfitting.

\emph{[Later Note:  If you have 2 data points and fit a parabola it would be hard work to find \textbf{any} new data point that displayed predictive estimability.  But if your new data point was close to one of the original data points
 (which causes near predictive estimability) you can still hope to get a good prediction.]}

\subsection{Best Linear Prediction}
When $(y,z')'$ is a random vector, the best linear predictor (BLP) of $y$ from $z$ is the ``linear conditional expectation''
    $$\Ehat(y|z) \equiv \mu_y + (z-\mu_z)'\beta_*,$$
where $\E(y)=\mu_y$, $\E(z)=\mu_z$, and $\beta_*$ is any solution to $\Sigma_{zz}\beta_*=\Sigma_{zy}$, e.g. Christensen (2020, Section 6.3).
This result holds whether or not $\Sigma_{zz}$ is nonsingular.

When the $(y_i,z_i')'$s form a random sample from some population, the obvious thing to do is to replace the unknown first and second order moments with their unbiased
estimates:  $\muhat_y=\ybar_\cdot$, $\muhat_z=\zbar_\cdot$, $\Sigmahat_{zz} = S_{zz}$, and $\Sigmahat_{zy} = S_{zy}$ which
leads to taking $\betahat_*$ to be any solution to the normal equations,
    $$ S_{zz}\betahat_* = S_{zy},$$
and the natural estimate of the BLP is
    $$\yhat \equiv \ybar + (z-\zbar)'\betahat_*.$$
These are precisely the predictions associated with the OLS estimates for the
reparameterized partitioned model (3.2).

Similar to the fact that for linear models,
estimable functions depend only on $N_*\beta_* \in C(S_{zz})$ and not on
$(I-N_*)\beta_* \perp C(S_{zz})$, BLPs depend only on a certain part of $\beta_*$.
Decompose $\beta_*=\beta_1+\beta_2$ with $\beta_1 \in C(\Sigma_{zz})$ and $\beta_2 \perp C(\Sigma_{zz})$.
Prediction depends only on $\beta_1$.
To see this, as in Christensen (2020, Lemma 1.3.5)  we know that for some random vector $b$ we have
$$\Pr\left[(z-\mu_z)=\Sigma_{zz}b \right] = 1.$$
It then follows that with probability one,
\begin{equation}
(z-\mu_z)'\beta_*= b'\Sigma_{zz}\beta_*= b'\Sigma_{zz}(\beta_1 +\beta_2)=b'\Sigma_{zz}\beta_1=(z-\mu_z)'\beta_1.
\end{equation}

Since $(z-\mu_z) \in C(\Sigma_{zz})$ a.s., when sampling future observations $(y_{fi},z_{fi}')'$
from the same population as the $(y_{i},z_{i}')'$s, we might hope that it should be nearly true that $(z_{fi}-\zbar_\cdot) \in C(S_{zz})$
which means that predictive estimability is nearly true and we can make good predictions.
In standard regression with $p \leq n$ it
is generally assumed that $r(\Sigma_{zz}) = p-1$ so $C(\Sigma_{zz})=\mathbf{R}^{p-1}$.
If the $z_i$s have an absolutely continuous distribution (which requires $\Sigma_{zz}$ to be nonsingular), then with probability one
$r(S_{zz})=\min(p-1,n-1)$, cf.~Okamoto (1973), so if $p \leq n$, then $r(S_{zz})=p-1$ and
$C(S_{zz})=\mathbf{R}^{p-1}$, so predictive estimability holds.  Unfortunately, when $p>n$, with probability one
$r(S_{zz}) = n-1 < p-1$ so $C(S_{zz})$ is strictly contained within $\mathbf{R}^{p-1}=C(\Sigma_{zz})$
and when $p>>n$ the spaces will not even be close.

For $p>n$ one can simply replace $S_{zz}$ with a more appropriate estimate.
There is a wide literature on such methods for estimating $\Sigma_{zz}$,
cf.~Fan, Liao, and Liu (2016) or Lam (2020) but, as Lam points out, achieving consistent estimation requires
one to impose a model for $\Sigma_{zz}$.  If the model is true, the predictions might be good but, for example,
spectral shrinkers (implicitly) provide estimates of $\Sigma_{zz}$ (as defined here
they estimate the precision matrix $\Sigma_{zz}^{-1}$) and, as we have seen,
even spectral shrinkers that give nonsingular estimates of a nonsingular $\Sigma_{zz}$
(like that associated with ridge regression)
do not solve the predictive estimability problem for spectral shrinkage estimates.
Spectral shrinkage estimates remain estimates within $C(S_{zz})$,
so they still estimate $N_*\beta_*$ rather than all of $\beta_*$.  In fact, the same can be said for any estimate
of $\Sigma_{zz}$ that takes the form $\Sigmatilde_{zz} = VDV'$ for \emph{any} diagonal matrix $D$.

\emph{[Later Note:  Any time $r(S_{zz}) = r(\Sigma_{zz})$ we should have predictive estimability with 
probability one.  With probability one, $(z_i-\mu_z) \in C(\Sigma_{zz})$, so 
$(\bar{z}_\cdot-\mu_z) \in C(\Sigma_{zz})$ and $(z_i-\bar{z}_\cdot) \in C(\Sigma_{zz})$, hence 
$C(S_{zz}) \subset C(\Sigma_{zz})$.  Also, with the ranks being equal, $(z_{fi}-\mu_z) \in C(\Sigma_{zz})$, 
so $(z_{fi}-\bar{z}_\cdot) \in C(\Sigma_{zz}) =C(S_{zz})$.]}

In the next section we argue that for complicated models with large $p$, the covariance matrix
$\Sigma_{zz}$ can sometimes be well approximated by a matrix of much lower rank than $p-1$.
This in turn suggests that it should be nearly true that $(z_{fi}-\zbar_\cdot) \in C(S_{zz})$,
hence ensuring that nonestimability is a a minor problem and that predictions should typically work well.

All of this requires us to sample the $(y_{fi},z_{fi}')'$s from the same population as the $(y_{i},z_{i}')'$s.
While in practical prediction problems this may be a questionable assumption, when
randomly dividing a complete set of data into training and test data we ensure that the
same model will apply to both parts.
While this does not ensure that $C(X_f')\subset C(X')$, i.e., predictive estimability, it does assure
that for fixed $p$ with large $n$ and $n_f$,
$$S_{zz} \rightarrow \Sigma_{zz} \leftarrow S_{fzz},$$
in probability.  Here $S_{fzz}\equiv\frac{1}{n_f-1}Z_f'[I_{n_f} - (1/n_f)J_{n_f}^{n_f}]Z_f$.
Our hope for predictive estimability is that
    $$C(\Sigma_{zz}) \doteq C(S_{fzz}) \doteq C([Z_f- J_{n_f}\zbar_\cdot']'[Z_f- J_{n_f}\zbar_\cdot']) = C([Z_f- J_{n_f}\zbar_\cdot']') \subset C(S_{zz}) \doteq C(\Sigma_{zz}) $$
The second approximate equality stems from having $\zbar_{f\cdot}$ replaced by $\zbar_{\cdot}$ in computing the
estimated covariance matrix of the future predictors.
The subset relation is what we need to have in order to get predictive estimability.
Unfortunately, for $p>n>n_f$ these approximations are unconvincing.

\subsection{Recapitulation}
To recap, if model (1.1) is a good model, even when $p>n$ we have tools to get good estimates of $N_*\beta_*$.
Any time that predictive estimability holds in model (3.1), that is, if
     $$C([Z_f- J_{n_f}\zbar_\cdot']'[Z_f-J_{n_f}\zbar_\cdot']) = C([Z_f-J_{n_f}\zbar_\cdot']') \subset C(S_{zz}),$$
good estimation of $N_*\beta_*$ is sufficient for good prediction.  With  $p>n$ we can probably only hope
that predictive estimability will be approximately true in which case we can hope that
$[Z_f-\zbar_\cdot' J_{n_f}](I-N_*)\beta_*$ is close to 0 so that it does not have a deleterious affect on
predictions.  We can easily check when predictive estimability is approximately true by doing a multivariate
regression of $[Z_f- J_{n_f}\zbar_\cdot']'$ on $Z'[I-(1/n)J_n^n]$.

When the predictive model (3.1) is sampled from the same population as (1.1), the fact that $C([Z_f- J_{n_f}\mu_z']') \subset C(\Sigma_{zz})$ with probability one gives us hope that  $C([Z_f- J_{n_f}\zbar_\cdot']') \subset C(S_{zz})$ is approximately true so that predictive
estimability is approximately true.  When both $n$ and $n_f$ are large, sample estimates from each model should be close to their population equivalents, so not only does
$S_{zz}$ approximate $\Sigma_{zz}$ but so do both $S_{fzz}$ and $(1/n_f)[Z_f- J_{n_f}\zbar_\cdot']'[Z_f-J_{n_f}\zbar_\cdot']$.
Again this gives us a basis to hope that predictive estimability may be approximately true
but for $p>n$ these approximations are complicated by rank deficiencies in the estimators as illustrated earlier.
%In the next section we argue that $\Sigma_{zz}$ itself should be modeled with rank deficiencies, thus ameliorating the rank deficiencies of the estimates.
As we will see in the next subsection, anytime $r(\Sigma_{zz}) <<n$, predictive estimability should hold approximately.

\subsection{Reduced Rank Covariance Matrices}

Suppose $r(\Sigma_{zz})=t-1<p-1$.  Under this assumption and random sampling of the $z_i$s,
the matrix $X$ in Section~1 has $r(X)=t$ (with probability 1) as assumed earlier.
A plausible model for this phenomenon is
the existence of a random $t-1$ vector $w$ with
absolutely continuous distribution, $\E(w)=0$, nonsingular $\Cov(w)=\Sigma_{ww}$, and a $p-1 \times t-1$
fixed matrix $R$ of rank $t-1$ with $z-\mu_z=Rw$.  Clearly
    $$\Sigma_{zz}= R\Sigma_{ww}R'.$$
Rewriting our independent data as $z_i-\mu_z=Rw_i$, leads to writing
    $$ Z = WR'+J_n \mu_z'.$$
Then
    $$S_{zz} = \frac{1}{n-1} Z'[I-(1/n)J_n^n]Z = \frac{1}{n-1} RW'[I-(1/n)J_n^n]WR' = R S_{ww}R' .$$
Regardless of the size of $p$, if $n$ is large relative to $t-1$, then $S_{ww}$ will be a good estimate of $\Sigma_{ww}$, so with $R$ fixed,
even if $R$ is unknown, $S_{zz}$ will be a good estimate of $\Sigma_{zz}$.

Since $(z-\mu_z) \in C(\Sigma_{zz})$ a.s., when sampling future observations $(y_{fi},z_{fi}')'$
from the same population as the $(y_{i},z_{i}')'$s, we can now expect it to be nearly true that
$(z_{fi}-\zbar_\cdot) \in C(S_{zz})$, so that predictive estimability will nearly hold and
all our predictions should work well.  But this only holds when $n$ is substantially larger than $t-1$.

\subsection{Spiked Covariance Matrices}
Our hope is that the argument of the previous subsection remains true when the reduced rank covariance
structure is only approximately true.
In practice we anticipate observing $r(S_{zz})=n-1<p-1$ but Section 4 argues that rank $t-1<<n$
reduced rank models are often reasonable approximations when $p>n$.  Spiked covariance
models for $\Sigma_{zz}$ provide rank $p-1$ models that approximate the behavior of
rank $t-1$ covariance structures.  Spiked models have been the subject of considerable research in
an asymptotic scenario with $p/n \rightarrow \gamma$.  Here $t$ is no longer $r(X)$.
Here most often $r(X)$ is either $n$ or $p$ depending on whether $\gamma >1$ ($p>n$ ) or $\gamma<1$ ($p<n$).
The standard asymptotic setting where everything works like we want is $\gamma=0$.
(All the theoretical work I have seen uses simplifying assumptions rather than the partitioned model.)

The original spike model for the eigenvalues
of $\Sigma_{zz}$ has a finite number $t-1$ of large eigenvalues with the remaining eigenvalues being constant.
Generalizations of this model focus on weakening the assumption that the remaining eigenvalues are constant.
One more realistic model is the generalized spike model of Bai and Yao (2012).
While this model is quite complicated and Dey and Lee (2019) comment on the limited work
performed on this generalized spike model, Dey and Lee have illustrations that seem to
suggest that the model is consistent with $t-1$ large eigenvalues and then exponential decay of
the the remaining eigenvalues.  (This comment is for fixed $n$ and $p$.  The asymptotics often require the
spike eigenvalues to grow with $p$ which complicates the meaning of exponential decay of
the nonspike eigenvalues.)  Jung (2022) considers a somewhat different generalization of the spike model
that also seems consistent with exponential decay for fixed $n$ and $p$.  One reason to emphasize exponential
decay of the eigenvalues is that Thibeault, Allard, and Desrosiers (2024) have found that spikes together
with exponential decay of the remaining eigenvalues characterize
wide classes of complicated models across various disciplines.

The main result of the spike covariance theory is that for $\gamma>0$ the first $t-1$ eigenvectors of $S_{zz}$
converge to biased versions of the first $t-1$ eigenvectors of $\Sigma_{zz}$.  This in turn implies that the
first sample principal components converge to biased versions of the
corresponding population principal components.  Changing his notation to agree with ours,
Jung (2022) states, ``the first $t-1$ sample and prediction [principal component] scores are
comparable to the true scores. The asymptotic relation tells that for large $p$,
the first $t-1$ sample scores ... converge to the true scores ..., \emph{uniformly
rotated and scaled for all data points}.''  (My italics.)
The point of principal component regression is reducing dimension while retaining as much as possible of the
information in the $z$ vector.  As discussed in Christensen (2019, Section~14.1),
any nonsingular linear transformation of a group of principal components contains
the same (linear) information about $z$ as the original principal components.
Since rotation and scaling are linear transformations,
for the purpose of doing principal component regression on $r \leq t-1$ principal components,
the asymptotic bias of the sample principal components relative to the population principal components
is irrelevant.

Although the work on the generalized spike models seems to focus on
the value of the eigenvectors associated with the spikes,
equally important for our purposes is that we not be leaving valuable information
on the table by ignoring the nonspike eigenvalues.
Jung, Lee, and Ahn (2018) indicate that for the original spike model
``the remaining estimated [principal component] scores are mostly accumulated noise''.
If so, they should be of no use in predicting $y$.  Other work on spike models also suggests that the nonspike
sample eigenvectors are uninformative about the corresponding population eigenvectors.
(Specifically, that the sample and population eigenvectors are asymptotically orthogonal.)

Write a SVD
    $$\Sigma_{zz} = \Vtilde D(\stilde_j)\Vtilde'; \qquad \Vtilde=[\Vtilde_1, \ldots, \Vtilde_{p-1}]$$
As discussed, for example, in Christensen (2019, Section 14.1),
$\sum_{j=r+1}^{p-1} \stilde_j/\sum_{j=1}^{p-1} \stilde_j$ measures the relative
amount of (linear) information in $z$ lost by using only the first $r$ population principal components.
Under exponential decay, this can be a relatively small amount.  Under a spike model the amount of
information retained from considering only the spike principal components is
$\sum_{j=1}^{t-1} \stilde_j/\sum_{j=1}^{p-1} \stilde_j$ which can be the vast majority of the information.
But unfortunately, there is no way to guarantee that the little information given up on $z$
by using only the spike components
is not the most relevant information for prediction.  As indicated earlier,
one can argue that the information being given up on $z$ is not reliable.  With training and test data
we can try our procedures and see if they work.  When performing nonparametric multiple regression using linear
combinations of spanning (or basis) functions, in theory any choice of spanning functions should work
(polynomials, trig functions, wavelets, etc.), but in practice
different basis functions work better or worse on different data and they have different nonestimable
parameters $(I-N_*)\beta_*$.

Again, the data only allow us to estimate $N_*\beta_*$, so any time $(z_{fi}-\zbar_\cdot)'(I-N_*)\beta_* \neq 0$
our predictions can be off.  In the $p>n$ linear model, we can never guarantee that we will do
a good job of predicting in the absence of predictive
estimability because we cannot guarantee that $(z_{fi}-\zbar_\cdot)'(I-N_*)\beta_*$ will be small.
In particular problems it may work out.  Similarly, best linear prediction from $z$ is equivalent to best
linear prediction from all $p-1$ population principal components and
there is no guarantee that the population principal components with small eigenvalues are not important
for predicting $y$.  However, if we are incapable of estimating the small eigenvalue
population principal components, it becomes irrelevant whether they are useful for predicting $y$.
We can only use as predictors the sample principal components that
give us meaningful information about the structure of $\Sigma_{zz}$.  If that is enough to give good
predictions, that is fantastic.

Similar to the previous subsection, if the spike model is close to being a rank $t-1<<n$ model,
the linear model should display approximate predictive estimability.
If, as Jung's (2022) Theorem 1 asymptotics suggest,
$C(\Vtilde_{t-1})\doteq C(V_{t-1}) \subset C(S_{zz}) = C(N_*)$
where $V_{t-1}$ and $\Vtilde_{t-1}$ now contain the first $t-1$ columns of $V$ and $\Vtilde$, then
$ C(S_{zz})^\perp \subset C(V_{t-1})^\perp \doteq C(\Vtilde_{t-1})^\perp$, and one can argue that
$(z-\zbar)'(I-N_*)\doteq 0$ because that occurs if and only if $\|(I-N_*)(z-\zbar)\|^2\doteq 0$ and,
using the standard result on the expected value of a quadratic form,
$$\displaylines{\quad \|(I-N_*)(z-\zbar)\|^2 \leq \| (I-V_{t-1}V_{t-1}')(z-\mu)\|^2 \doteq \E \| (I-\Vtilde_{t-1}\Vtilde_{t-1} ')(z-\mu)\|^2  \hfill \cr \hfill
\mbox{} = \E (z-\mu)'(I-\Vtilde_{t-1}\Vtilde_{t-1} ')(z-\mu)  = \tr\left[(I-\Vtilde_{t-1}\Vtilde_{t-1} ')\Vtilde D(\stilde)\Vtilde\right] = \sum_{j=t}^{p-1} \stilde_j \doteq 0 ,\quad}$$
where the last approximate equality is the definition of what it means for the spike model to be close
to a rank $t-1$ model.  Of course it is actually easy to check approximate predictive estimability
rather than relying on this argument.  It also bears repeating that we will never know $t$.

\setcounter{equation}{0}
\section{A Model for Overfitting}

One common procedure leading to linear models with more parameters than observations is performing
nonparametric multiple regression.  For example, if you observe the vector $x'=(1,z')$, fitting an $r-1$ degree
interactive polynomial model in each component of $z$ leads to the linear model
    $$\E(y|z)=\sum_{k_1=0}^{r-1} \cdots \sum_{k_{p-1}=0}^{r-1} \beta_{k_1 \ldots k_{p-1}}
z_1^{k_1}\cdots z_{p-1}^{k_{p-1}},$$
which is the sum of $r^{p-1}$ terms.  The same idea applies when fitting trig functions, splines, or
wavelets in which case, using definitions of functions $\phi_{h}$ appropriate for the application,
    $$\E(y|z)=\sum_{k_1=0}^{r-1} \cdots \sum_{k_{p-1}=0}^{r-1} \beta_{k_1 \ldots k_{p-1}}
\phi_{k_1}(z_1)\cdots\phi_{k_p}(z_{p-1}).
$$
Here most often $\phi_0 \equiv 1$.  The number of terms $r^{p-1}$ quickly gets out of hand.
If you measure just $p-1=5$ variables in $z$ and use $r=8$ functions $\phi_h$
to model the curve in each dimension, then
the total number of parameters in the constructed linear model is nearly 33,000.

In general we assume that there is some underlying $p-1$ vector of predictor variables $z$ and that for $z$
there is some vector valued function $\phi(z)$ taking values in  $\mathbf{R}^s$ for which
    \begin{equation} \E(y|z) = \beta_0 + \phi(z)'\beta_*.\end{equation}
In this model, our concern is estimation when $s+1>n$ rather than with $p>n$.
In this model we can plausibly have $p<<n$.  The key idea is that a limited number of actual
random variables in $z$ are driving a linear explanatory model based on $\phi(z)$ of very high dimension
and that this functional relationship drives the applicability of the covariance
models discussed in Subsections~3.3 and 3.4, wherein the roles of $t-1$ and $p-1$ are now being played by
$p-1$ and $s$, respectively.

As illustrated for nonparametric multiple regression,
we often have $s>>p$.  While $s$ may need to be large to get a good fitting model,
often one picks $s$ much larger than it needs to be to ensure that
the (unknown) relevant features of $\E(y|z)$
are being captured.  If model (4.1) involves
such overfitting it is incumbent upon us to adjust for that by using
some form of shrinkage estimate like: fitting reduced models (e.g., principal components with $r<n$),
penalized least squares, or spectral shrinkage estimates.

For nonparametric multiple regression, $z$ is observed and $\phi$ is known, so $\phi(z)$ is observed.
Moreover, for nonparametric multiple regression, we know that by taking $s$
sufficiently large with appropriate $\phi_{h}$s, model (4.1) \emph{will} be approximately correct.
However, in general, we do not need to see $z$ or know $\phi$ we only need to be able to observe $\phi(z)$.
In a general problem with a large number $s$ of observable predictors,
we can imagine that there are $p-1$ unobserved underlying
variables $z$ that drive the entire problem and that our $s$ observations $\phi(z)$ are unknown
functions of those unobserved underlying variables.

Write $\E[\phi(z)] \equiv \mu_\phi$, $\Cov[\phi(z)] \equiv \Sigma_{\phi \phi}$ and $\Cov[\phi(z),y]\equiv \Sigma_{\phi y}$, then the best linear predictor is
$$\Ehat[y|\phi(z)] = \mu_y + [\phi(z)-\mu_\phi]'\beta_*$$
where $\beta_*$ is any solution of $\Sigma_{\phi \phi}\beta_* = \Sigma_{\phi y}$.
A first order Taylor's expansion of $\phi(z)$ provides a linear (actually affine) approximation to $\phi(z)$ in terms of $z$,
\begin{equation}
 \phi(z) \doteq \phi(\mu_z) + \left[\mathbf{d}_z \phi(\mu_z) \right] (z-\mu_z).
\end{equation}
While we will continue to use this Taylor's approximation, the fundamental idea simply
requires a reasonable linear approximation of $\phi(z)$ based on $z$.
Based on the linear approximation,
$$\Sigma_{\phi \phi} \doteq \left[\mathbf{d}_z \phi(\mu_z) \right] \Sigma_{zz} \left[\mathbf{d}_z \phi(\mu_z) \right]'.$$
If (4.2), or any other similar linear approximation, happens to hold \emph{exactly}, then the reduced rank
covariance structure of Subsection~3.3 applies.  If (4.2), or any other similar linear approximation, applies,
then the spike covariance structure of Subsection~3.4 seems plausible.
Again, although not directly relevant to this problem, Thibeault et al.~(2024)
found the spike model with exponential decay applicable to a wide variety of complex systems.

That $\Sigma_{\phi \phi}$ and $S_{\phi \phi}$ can be very nearly singular is well-known.
Even in just one dimension with $p<n$, fitting high degree polynomials are notorious for
their numerical difficulties due to the collinearity in $X'X$ and $Z'[I-(1/n)J_n^n]Z$.
Hardly anyone fits $y_i = \beta_0 + \beta_1 z_i + \cdots + \beta_5 z_i^5 + \vep_i$
because of high correlations among the predictors.  One is far more likely to fit
$y_i = \gamma_0 + \gamma_1 (z_i-\zbar_\cdot) + \cdots + \gamma_4 (z_i-\zbar_\cdot)^4
+ \beta_5 (z_i-\zbar_\cdot)^5 + \vep_i$
but even then collinearity problems often persist (leading some to fit orthogonal polynomials).
The point is that when fitting the raw or even the mean corrected polynomial
for this somewhat complex model, $\Sigma_{\phi \phi}$ and $S_{\phi \phi}$ will be very nearly singular.
Unfortunately, as we will see by example in the next section, this near singularity may not be sufficient to ensure
that $t-1<<n$ as required for a strong hope of approximate predictive estimability.

\section{Double Descent}

Double descent is a phenomenon associated with the prediction errors from
fitting increasingly large models to a fixed set of data.
We revert to the notation of Section~1 rather than Section~4.
The idea in our linear models is that $n$ is fixed and $p$ is increasing.
The phenomenon is the empirical observation that often when $p$ is small we under fit the data, as $p$
increases we get increasingly better predictions but as $p$ approaches $n$ we begin to overfit the model
and get worse results to the point that when $p=n$ we are often interpolating the data ($\yhat_i=y_i$)
and getting poor predictions.
The double in double descent is the observation that as $p$ gets bigger than $n$, we again see improved
predictive performance.

The unexpressed catch is that as $p$ increases, we need to specify the changing model matrix $X$,
which we now refer to as $X_p$.   To observe double descent, you need some dross among the sequence of models.

Double descent is really about under fitting, rather than over fitting.  The idea is that for $p<n$,
the $X_p$ models have a limited capability for modeling $\E(y|x)$ and this capability
is at first fully exploited and then subjected to overfitting as $p \rightarrow n$.  The second descent,
when $p>n$, comes from using increasingly complex $X_p$ that actually give good approximations to $\E(y|x)$.

For example, we know that standard methods for nonparametric multiple regression
can approximate any continuous $\E(y|x)$ but that getting a good approximation often
requires $p>n$, so, in the absence of predictive estimability,
we cannot be sure that our parameter estimates will yield good predictions.
Of course when test data are available, we can actually see which, if any, specific procedures work well.
Another problem is that we do not know the structure of $\E(y|x)$, so we do not know how large to make $p$
in order to catch the appropriate structure, so we often pick a $p$ that overfits.
Of course this is not an issue when looking specifically at the double descent phenomenon,
wherein one fits each $X_p$ model in the sequence.
Even with $p>n$ we would be inclined to stop fitting when the predictive fits started getting worse,
although there is no reason to believe that a phenomenon of triple descent could not be possible.

Hastie et al.~(2022) provide a theoretical discussion of this phenomenon with numerous references.
Here we merely illustrate the double descent phenomenon with a simple simulated example.
Online supplemental material includes a more extensive discussion of our example and
our actual code.  We now revert to the notation from Section~4 with $p-1$ original predictor variables and nonparametric multiple
regression models with a total of $s+1$ predictors.

We (Mohammad Hattab and I) generated data
according to a multidimensional noninteractive symmetric fourth degree polynomial $\E(y|z)=\sum_{j=1}^{p-1} z_j^4-2z_j^2$.
%Because we are fitting polynomials, we revert to the notation of Section~4.
Figure 1 shows the form of the regression function in each variable.
The idea is to fit a sequence of polynomial models using least squares, that the intercept and linear terms will be of little use, the quadratic term
should have some imperfect utility in explaining the tail behavior,
the number of sample observations will be chosen to cause a cubic polynomial to interpolate the training
data, but we will continue to fit correct fourth and fifth degree polynomials with $s+1>n$ using Moore-Penrose least squares estimates
(minimum norm least squares estimates).

From Figure 1, if you sample predictor values primarily from the center of this curve, say between $-1.5$ and $1.5$,
a flat line does a remarkably good job of prediction
and fitting a quadratic will be of little help.
Many distributions for the predictor variables that are symmetric with a mode at 0, like (multivariate) normals and $t(df)$ distributions, fit this pattern.
You have to have substantial interest in values out near $\pm 2$ before fitting a quadratic starts to help.
To achieve this we initially used independent $U(-2,2)$ distributions
for the predictor variables and finally settled on sampling correlated data with  $U(-2,2)$ marginal distributions
for both the training and test data.

\begin{figure}\begin{center}
\includegraphics[width=4.5in,height=3in]{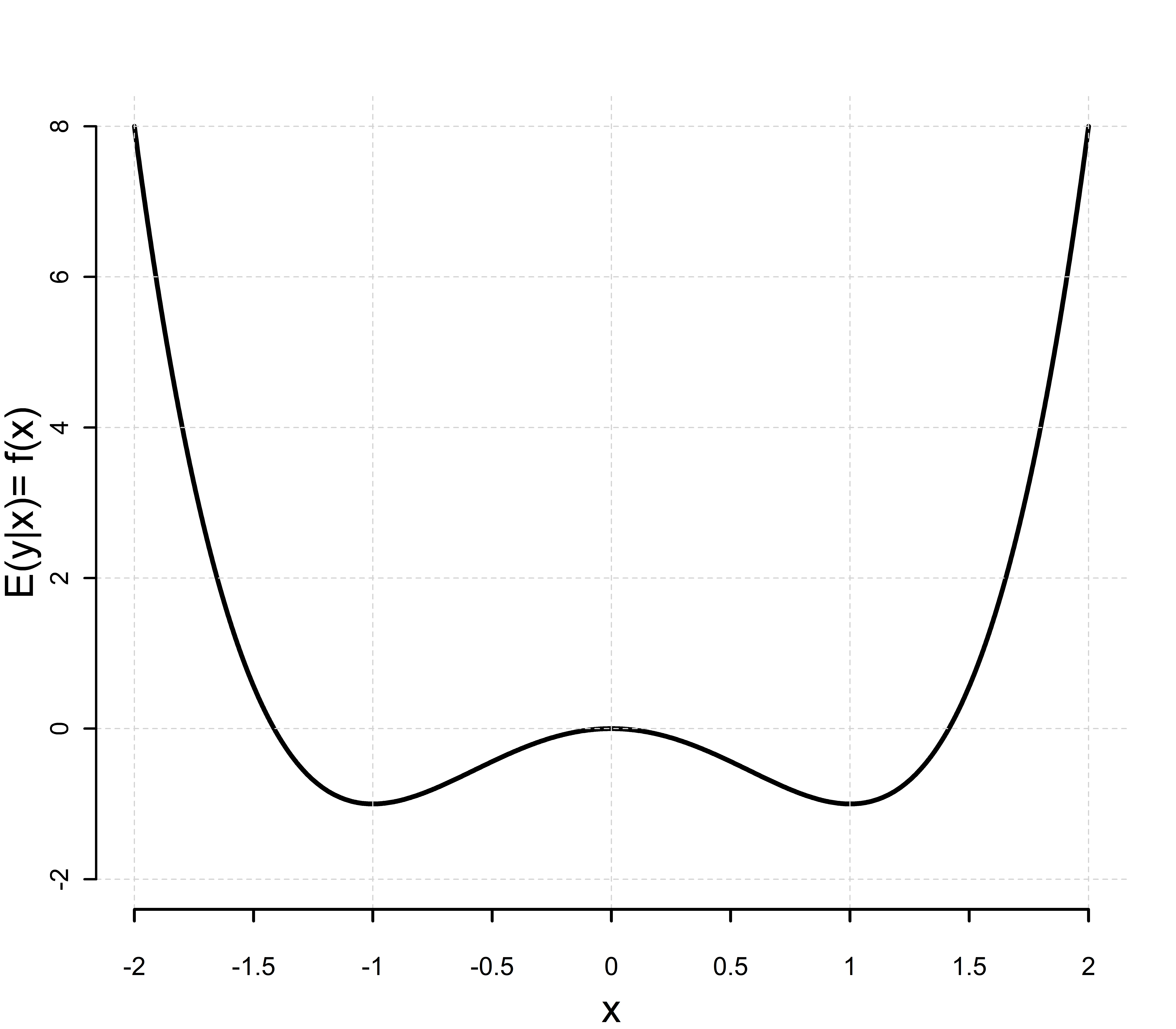}                  %{c:/e-drive/Books/LINMOD23/plots/fig1-1.eps}
\caption{Functional form of $E(y|x_j)=x_j^4-2x_j^2$.}
\end{center}\end{figure}

The sequence of models to be fitted involve no interactions and incorporate an intercept only model, a model that adds all linear
terms in the various dimensions, then all quadratics, then cubics, then quartics, then quintics,
so our sequence involves only six models and our sequence of models have numbers of parameters $s+1=1,p,1+2(p-1),\ldots,1+5(p-1)$.

We trained on $n=1+3(p-1)$ observations so that $n=s+1$ when fitting the cubic polynomial and we took a test sample size of $n_f=101$.
With truth being a symmetric 4th degree, fitting the linear and cubic terms should be useless,
the quadratic terms should model useful aspects of the underlying truth and reduce both training error and
prediction error relative to the linear terms.  The cubic terms
overfit and interpolate the data,
so reduce training error to 0
but raise the prediction error by incorporating and estimating useless terms.  With $p>n$ we
incorporate the true quartic features of the model.   The 4th degree model
captures enough of the true model to improve prediction beyond the quadratic model.
The quintic model again constitutes overfitting.

From $r=1,\ldots, 1000$ different $Y$ vectors and for each model we computed a simulation estimate of the squared deviation between our predictions and the best
predictions, i.e., we simulated $\E_{x_f,Y}[\E(y_f|x_f)-x_f'\betahat(Y)]^2$ using
 $$PMSE \equiv \frac{1}{1000} \sum_{r=1}^{1000} \left\{\frac{1}{n_f} \sum_{i=1}^{n_f} \left[\E(y_{fi}|x_{fi})- x_{fi}'\betahat_r \right]^2 \right\}.$$
We have a relatively small sample of predictions, $n_f=101$, but the goal is really to get accurate predictions rather than an accurate estimate of
the expected squared predictive estimation error.
We also computed a simulation estimate of the mean squared prediction biases,
$$Bias^2 = \frac{1}{n_f} \sum_{i=1}^{n_f} \left[\E(y_{fi}|x_{fi}) - \frac{1}{1000} \sum_{r=1}^{1000} x_{fi}'\betahat_r\right]^2.$$
When we are fitting a correct linear model with $s+1>n$ (in our simulations, either the quartic or quintic models), this becomes
$$Bias^2 = \frac{1}{n_f} \sum_{i=1}^{n_f} \left(x_{fi}'\beta - \frac{1}{1000} \sum_{r=1}^{1000} x_{fi}'\betahat_r\right)^2 \doteq \frac{1}{n_f} \sum_{i=1}^{n_f} \left(x_{fi}'\beta - x_{fi}'N\beta\right)^2 = \frac{1}{n_f} \sum_{i=1}^{n_f} \left[x_{fi}'(I-N)\beta\right]^2,$$
so when there are fewer observations than model terms, this squared bias term involves the parts of the Best Predictor that we are incapable of estimating.

Table 1 contains simulation results for fitting polynomial models with 50 predictors,
on the left with the number of observations equal to the number of parameters in the cubic model
and on the right with twice that number of observations minus 1.
With different random selections of predictor variables the numbers in these tables can change a fair amount but the patterns remain pretty consistent.
The table on the left displays the classic pattern of double descent with a big drop-off in PMSE for fitting the quadratic model, a huge increase for
fitting the cubic model that interpolates the data, a huge drop-off for fitting the correct quartic model even though there are fewer observations than predictors,
and then a moderate increase due to overfitting with the quintic model.  Most of the prediction error is due to estimation bias except when fitting the
optimal quartic model in which case bias is still substantial.  Of course the order of magnitude of the bias problem for the correct quintic model is much
less than the bias for the incorrect models.  The bias tells us that neither of the two correct models has $\beta$ all that close to $N\beta$,
so, although the correct models greatly improve predictions, the predictions leave much to be desired.

\begin{table}[!htbp]
\caption{Error in Estimating Best Predictors:  Polynomial models with $p-1=50$, $n_f=101$.}
\centering
\begin{tabular}{rrr|r|rrr}   \hline
\multicolumn{3}{c|}{$n=151$} &     &  \multicolumn{3}{|c}{$n=301$}        \\
Model& $PMSE$ & $Bias^2$ &  $s+1$ &  Model & $PMSE$ & $Bias^2$       \\  \hline
$M_0$&  3905.59  &    3905.59  &    1 &  $M_0$ & 4254.91  & 4254.91  \\
$M_1$&  4800.50  &    4799.97  &   51 &  $M_1$ & 5384.33  & 5384.12  \\
$M_2$&   364.01  &     361.91  &  101 &  $M_2$ &  260.05  &  259.55  \\
$M_3$& 82360.71  &   80054.88  &  151 &  $M_3$ &  483.90  &  482.80  \\
$M_4$&    12.30  &       6.81  &  201 &  $M_4$ &    2.32  &  0.0021  \\
$M_5$&    59.84  &      53.82  &  251 &  $M_5$ &    6.55  &  0.0060  \\
   \hline
\end{tabular}
\end{table}

The table on the right displays traditional regressions with more observations than parameters.  With the larger sample size, estimability
holds for all predictions.  The pattern in PMSE remains the same for models up to the quadratic
but the cubic model is no longer an interpolation model and is no longer severely overfitted.  The cubic term is still useless, so it constitutes overfitting relative
to the useful but imperfect quadratic model.  The quality of predictions
for the correct quartic and quintic models is much better with the larger sample size because the parameter estimates are now theoretically unbiased.
The quintic model is overfitted so behaves somewhat worse.  As before, for models of degree less than 4, most of the prediction error is due to estimation bias.

For $n=151$, of the 200 eigenvalues in $S_{zz}$ for the optimal fourth degree model,
the largest 49 accounted for 90\% of the total and the largest 86 accounted for 99\% of the total.  While the covariance matrix
is very nearly singular, it does not have $t-1<<n$ as discussed in Section~3 for getting everything close to predictive estimability.
For $n=301$, of the 200 eigenvalues in $S_{zz}$ for the optimal fourth degree model,
the largest 59 accounted for 90\% of the total and the largest 96 accounted for 99\% of the total.

Finally, we also examined much larger models with 500 predictor variables.
Again, it takes roughly half of the eigenvalues to account for 99\% of the variability in the quartic model.
The results presented in Table~2 are qualitatively comparable to those in Table~1.
The numbers being reported are averages, so the larger numbers for the incorrect models with 500 predictors rather than 50 predictors,
suggest that wrong models are a bigger problem with more predictors.
It seems even more clear from the larger problem in Table~2 that \emph{fitting a appropriate model is far more
important than having enough observations to fit that model using traditional methods}, as desirable as it is to have additional observations.

\begin{table}[!htbp]
\caption{Error in Estimating Best Predictors:  Polynomial models with $p-1=500$, $n_f=101$.}
\centering
\begin{tabular}{rrr|r|rrr}   \hline
\multicolumn{3}{c|}{$n=1501$} &     &  \multicolumn{3}{|c}{$n=3001$}        \\
Model& $PMSE$ & $Bias^2$ &  $s+1$ &  Model & $PMSE$ & $Bias^2$       \\  \hline
$M_0$ & 266827.6     &       266827.6 &    1  & $M_0$   &  275783   & 275783      \\
$M_1$ & 333943.4     &       333942.9 &  501  & $M_1$   &  293307   & 293307      \\
$M_2$ &  20047.2     &       20045.3  & 1001  & $M_2$   &  10027    & 10026       \\
$M_3$ & 109707117.~~ &  109701539.~~  & 1501  & $M_3$   &  15293    & 15292      \\
$M_4$ &     12.3     &           7.1  & 2001  & $M_4$   &     2.2   & 0.0023   \\
$M_5$ &     46.6     &          41.1  & 2501  & $M_5$   &     5.6   & 0.0066   \\
   \hline
\end{tabular}
\end{table}

\section{Final Comments}

When $p>n$ we can always decompose $\beta$ into nontrivial components $N\beta \in C(X')$ and $(I-N)\beta \in  C(X')^\perp$.   $N\beta$ is identifiable and estimable and the minimum norm least squares estimate is unbiased for it.  $(I-N)\beta$ has no effect on the distribution of $Y$, so it is impossible to learn about it from the data.
The parameter $(I-N)\beta$ can only be informed \emph{a priori}.
Many popular estimation methods focus on estimating $N\beta$ as they provide estimates that only exist in $C(X')$.

In the models considered, the best predictor is always the expected value of the future observation to be predicted.  When these quantities are estimable, they are functions of $N\beta$ and standard results on
prediction and estimation apply.    For prediction means that are not estimable,
we have considered sampling models for the rows of $X$ that suggest that most future observations will
have means that are close to being estimable, so that there is hope (but no assurance) that the unidentifiable parameter $(I-N)\beta$ will have little effect on the best predictor.  Regardless of the sampling model, it is easy to check whether the mean of a future observation is close to being estimable.  When the overall data are randomly divided into training and test subsets, it is also easy to check whether $(I-N)\beta$
is playing a large role because that would cause poor predictions in the test data.

A simple way to proceed in these problems is to replace model (1.1) with the principal component model using
the principal components for all the positive eigenvalues of $S_{zz}$.  This model can be explored
with the usual array of reduced models and biased estimation techniques but is hindered by the fact that,
if there are no replications in the rows of $X$, typically it would interpolate the data and give $SSE=0$.

With $p>n$, typically there will be insufficient degrees of freedom for error to obtain a reliable estimate of $\sigma^2$ from model (1.1).  As discussed in Christensen (2024) [but eliminating his typos], if you have test data and predictive estimability, you can use  $(Y_f-X_f\betahat)'[I+X_f(X'X)^-X_f']^{-1}(Y_f-X_f\betahat)/n_f$ as an unbiased estimate of $\sigma^2$.  Under normality this has the usual relationship to a $\chi^2(n_f)$
distribution but will not be independent of typical mean squares for hypotheses regarding model (1.1).
Without predictive estimability, one could check to see if there are enough predictive cases
with estimable means to give an adequate estimate of variance or,
if that is not the case, one
could see if there are sufficient cases with either predictive estimability or near predictive estimability
to get a useful idea of the variance from this procedure.

Another option is that if sample principal components corresponding to
small but positive eigenvalues have no relation to their corresponding population principal components, they
could be dropped from the model and used for estimating error.  The result seems likely to be biased.
Alternatively, if, as one typically would, we have adjusted the $z$ variable to have a common scale,
we can probably drop the $n-t$ principal components with the smallest estimated regression coefficients.
In my (admittedly limited) experience, this is pretty much what the LASSO would do to the principal component model.

When the original model (1.1) is over fitted, the various biased estimation methods
may suggest reductions that can be made in the original model.  For example, if (1.1) is a 6th degree polynomial
in many predictor variables being fitted by principal components and the estimates of $\beta$ suggest
that the 5th and 6th degree terms are not important, then one could try fitting only a 4th degree polynomial.

Our discussion has focused on highly over-parameterized linear models.  Neural networks are examples of highly over-parameterized nonlinear regression models.

\section*{References}
\hangindent=2em
\begin{trivlist}

\item Bai, Z.D. and Yao, J. (2012). On sample eigenvalues in a generalized spiked population model, \emph{Journal of Multivariate Analysis}, \textbf{106}, 167-177.

    \item[] Christensen, R. (2015).   {\em  Analysis of Variance,
Design, and Regression:  Linear Modeling for Unbalanced Data}, Second Edition.  Chapman and Hall/CRC Pres, Boca Raton, FL.

    \item Christensen, Ronald (2019). \emph{Advanced Linear Modeling: Statistical Learning and
Dependent Data}, Third Edition. Springer-Verlag, New York.

    \item[] Christensen, Ronald (2020).  \emph{Plane Answers to Complex Questions:  The Theory of Linear Models}, Fifth Edition.  Springer, New York.

    \item Christensen, Ronald (2024). Comment on ``Forbidden Knowledge
and Specialized Training: A Versatile Solution for the Two Main Sources of Overfitting
in Linear Regression,'' by Rohlfs (2023), \emph{The American Statistician}, \textbf{78}, 131-133, DOI:
10.1080/00031305.2023.2277156

\item Dey, Rounak and Lee, Seunggeun (2019).  Asymptotic properties of principal component analysis and
shrinkage-bias adjustment under the generalized spiked
population model.  \emph{Journal of Multivariate Analysis}, \textbf{173}, 145-164.

\item Fan, J., Liao, Y., and Liu, H. (2016). An overview of the estimation of large
covariance and precision matrices. \emph{The Econometrics Journal}, 19(1), C1-C32.

\item Hastie, Trevor; Montanari, Andrea; Rosset, Saharon; and Tibshirani, Ryan J. (2022).
Surprises in highdimensional ridgeless least squares interpolation.
\emph{The Annals of Statistics}, \textbf{50}, 949-986.  \url{https://doi.org/10.1214/21-AOS2133}.

\item Hattab, Mohammad W.; Jackson, Charles S.; and Huerta, Gabriel (2019).
Analysis of climate sensitivity via high-dimensional principal component regression,
\emph{Communications in Statistics: Case Studies, Data Analysis and Applications},
DOI: 10.1080/23737484.2019.1670119

\item Hellton, K.H. and Thoresen, M.(2017).
When and why are principal component scores a good tool for visualizing high-dimensional data?
\emph{Scandinavian Journal of Statistics}, \textbf{44}, 581-816.

\item Jung, Sungkyu (2022). Adjusting systematic bias in high
dimensional principal component scores. \emph{Statistica Sinica}, \textbf{32}, 939-959.

\item Jung, S.; Lee, M.H.; and Ahn, J. (2018).
On the number of principal components in high dimensions.
\emph{Biometrika}, \textbf{105}, 389-402.

    \item Lam, Clifford (2020).  High-dimensional covariance matrix estimation.
\emph{WIREs Computational Statistics}, \textbf{12}(2).

\item Nosedal-Sanchez, A.; Storlie, C.B.; Lee, T.C.M.; Christensen, R. (2012). Reproducing
kernel Hilbert spaces for penalized regression: A tutorial. \emph{The American
Statistician}, \textbf{66}, 50-60.

\item Okamoto, M. (1973). Distinctness of the eigenvalues of a quadratic form in a multivariate sample. \emph{Annals of Statistics}, \textbf{1}, 763-765.

%Okamoto, M. and Kanazawa, M. (1968). Minimization of eigenvalues of a matrix and optimality of principal components. \emph{Annals of Mathematical Statistics}, \textbf{39}, 859-863.

\item Richards, Dominic; Dobriban, Edgar; and Rebeschini, Patrick (2021).
Comparing Classes of Estimators: When does Gradient Descent Beat Ridge Regression in Linear Models?
\emph{ArXiv}: 2108.11872.

\item Thibeault, V.; Allard, A.; and Desrosiers, P. (2024).  The low-rank hypothesis of complex systems.
\emph{Nature Physics}, \textbf{20}, 294-302.
\end{trivlist}

\section*{Appendix}

\bigskip
\textsc{Example:} \quad Find minimum norm and ridge estimates in overparameterized balanced one-way ANOVA
with $a$ groups. Let $J_r^c$ be an $r \times c$ matrix of 1s and $J_r \equiv J_r^1$.
$$X'X = \bmatrix{
aN & N & N & \cdots & N \cr
 N & N & 0 & \cdots &  0 \cr
 N & 0 & N &  & \vdots \cr
 \vdots & \vdots & & \ddots & 0 \cr
 N & 0 & \cdots & 0 & N  }  = \bmatrix{aN & N\, J_a' \cr N\,J_a & N\, I_a}.$$
The rank of this $(a+1)\times (a+1)$ matrix is $a$ and a vector in the orthogonal complement is defined by
$$\bmatrix{aN & N\, J_a' \cr N\,J_a & N\, I_a} \bmatrix{1 \cr -J_a}= 0.$$
With a basis for the orthogonal complement we can write the ppo onto $C(X')$ as
 $$N = I - \bmatrix{1 \cr -J_a}\left(\bmatrix{1 \cr -J_a}'\bmatrix{1 \cr -J_a}\right)^{-1}\bmatrix{1 \cr -J_a}' = I - \frac{1}{a+1}\bmatrix{ 1 & -J_a' \cr -J_a & J_a^a}$$
and, since $\betahat = [0,\ybar_{1 \cdot}, \ldots, \ybar_{a \cdot}]'$ is an OLS estimate, the unique minimum norm estimate is
$$N \betahat = \bmatrix{ \frac{a}{a+1}\ybar_{\cdot \cdot} \cr \ybar_{1 \cdot} - \frac{a}{a+1}\ybar_{\cdot \cdot} \cr \vdots \cr  \ybar_{a \cdot} - \frac{a}{a+1}\ybar_{\cdot \cdot} }.$$
Other choices for OLS given earlier yield the same result.

\bigskip

Ridge regression estimates require
 $$X'X+(\lambda N) I= N \bmatrix{a+\lambda & J_a' \cr J_a & (1+\lambda) I_a},$$
where, for convenience, the ridge parameter is $\lambda N$ rather than $\lambda $.
Using the standard formula for the inverse of a partitioned matrix (e.g., Christensen, 2020, Exercise B.21),
$$ \frac{1}{N} \bmatrix{a+\lambda & J_a' \cr J_a & (1+\lambda) I_a}^{-1} =  \frac{1}{N} \bmatrix{\frac{1+\lambda}{\lambda(a+1+\lambda)} & \frac{-1}{\lambda(a+1+\lambda)}J_a' \cr \frac{-1}{\lambda(a+1+\lambda)}J_a & \frac{1}{1+\lambda}\left(I + \frac{1}{\lambda(a+1+\lambda)}J_aJ_a'\right) }.$$

The ridge regression estimate is
$$\displaylines{\quad (X'X+\lambda N I)^{-1}X'Y  = \mbox{} \hfill \cr \hfill
 \bmatrix{\frac{1+\lambda}{\lambda(a+1+\lambda)} & \frac{-1}{\lambda(a+1+\lambda)}J_a' \cr \frac{-1}{\lambda(a+1+\lambda)}J_a & \frac{1}{1+\lambda}\left(I + \frac{1}{\lambda(a+1+\lambda)}J_aJ_a'\right) } \bmatrix{a\ybar_{\cdot \cdot} \cr \ybar_{1 \cdot} \cr \vdots \cr \ybar_{a \cdot}}
 = \bmatrix{\frac{a}{a+1+\lambda} \ybar_{\cdot \cdot}  \cr \frac{1}{1+\lambda}\left(\ybar_{1 \cdot} -  \frac{a}{a+1+\lambda} \ybar_{\cdot \cdot} \right)\cr \vdots \cr \frac{1}{1+\lambda} \left(\ybar_{a \cdot} -  \frac{a}{a+1+\lambda} \ybar_{\cdot \cdot} \right)}\quad }$$
Note that for $\lambda=0$ these are the minimum norm estimates.

Minimum norm OLS is minimizing $\| \beta \|^2$ subject to $\|Y-X\beta \|^2 = \|Y-X\betahat \|^2$ whereas ridge is minimizing
$\|Y-X\beta \|^2 $ subject to  $\| \beta \|^2=K$ where $K$ is determined by $\lambda$.

\bigskip
\textsc{Example:} \quad Find the minimum norm estimate in a balanced additive two-way ANOVA,
$$y_{ijk} = \mu + \alpha_i + \eta_j + \vep_{ijk}, \qquad  i=1,\ldots,a, j=1,\ldots,b, k=1,\ldots,N.$$
The model matrix can be written
$$X = \left\{ [J_a \otimes J_b \otimes J_N], [I_a \otimes J_b \otimes J_N], [J_a \otimes I_b \otimes J_N] \right\}$$
One well-known choice of least squares estimates is
$$ \betahat=\bmatrix{\muhat \cr \alphahat \cr \etahat} = \bmatrix{\ybar_{\cdot \cdot \cdot} \cr \Ybar_a - \ybar_{\cdot \cdot \cdot}\, J_a \cr \Ybar_{b} - \ybar_{\cdot \cdot \cdot}\, J_b },
\quad \hbox{where} \quad \Ybar_{a}= \bmatrix{\ybar_{1\cdot\cdot} \cr \ybar_{2\cdot \cdot} \cr \vdots \cr \ybar_{a \cdot\cdot} };
\quad  \Ybar_{b}= \bmatrix{\ybar_{\cdot 1 \cdot} \cr \ybar_{\cdot 2 \cdot} \cr \vdots \cr \ybar_{\cdot b \cdot} }.
$$
Note that $J_a'\Ybar_a=a\ybar_{\cdots}$ and $J_b'\Ybar_b=b\ybar_{\cdots}$.

We now go through the computations to find the minimum norm estimate.
$$X'X = \bmatrix{
abN   & J_a'bN & J_b'aN    \cr
bNJ_a & bNI_a  & J_a^b N  \cr
aNJ_b & J_b^aN & aNI_b     }$$
Define $R$ implicitly with $C(R)=C(X')^\perp$ via
$$ 0 = X'X \bmatrix{1 & 1 \cr -J_a & 0 \cr 0 & -J_b} = X'XR.$$

We now find $N\betahat$ where $N = I- R(R'R)^{-1}R'$.
$$R'R = \bmatrix{1+a & 1 \cr 1 & 1+b} ,\qquad (R'R)^{-1}= \frac{1}{a+b+ab}\bmatrix{1+b & -1 \cr -1 & 1+a},$$

$$(R'R)^{-1}R'=\frac{1}{a+b+ab}\bmatrix{b & -(1+b) J_a' & J_b' \cr a & J_a' & -(1+a)J_b' } ,$$ $$ (R'R)^{-1}R'\betahat = \frac{1}{a+b+ab}\bmatrix{b \ybar_{\cdots} \cr a \ybar_{\cdots}},$$

$$\betahat_m = N\betahat = \betahat - R \left\{ (R'R)^{-1}R'\betahat\right\} =  \bmatrix{\ybar_{\cdot \cdot \cdot} \cr \Ybar_a - \ybar_{\cdot \cdot \cdot}\, J_a \cr \Ybar_{b} - \ybar_{\cdot \cdot \cdot}\, J_b } - \frac{1}{a+b+ab}\bmatrix{b \ybar_{\cdots} + a \ybar_{\cdots} \cr - b \ybar_{\cdots} J_a \cr  - a \ybar_{\cdots}J_b }$$
or
$$\betahat_m = \bmatrix{\muhat_0 \cr \alphahat_0 \cr \etahat_0} =\bmatrix{\frac{ab}{a+b+ab}\ybar_{\cdot \cdot \cdot} \cr \Ybar_a - \frac{a+ab}{a+b+ab}\ybar_{\cdot \cdot \cdot}\, J_a \cr \Ybar_{b} - \frac{b+ab}{a+b+ab}\ybar_{\cdot \cdot \cdot}\, J_b } .$$
%Note that $\betahat_0$ remains a least squares estimate because $\yhat_{ijk} = \ybar_{i \cdot \cdot} + \ybar_{\cdot j \cdot} - \ybar_{\cdot \cdot \cdot} = \muhat_0 + \alphahat_{0i}+\etahat_{0j}$.

\end{document}